%% file: arxiv_main.tex
\documentclass{article} 

\input{math_commands.tex}

\usepackage{subcaption}
\usepackage{caption}
\usepackage{longtable}  

\usepackage{csquotes}
\usepackage[utf8]{inputenc} 
\usepackage[T1]{fontenc}    
\usepackage{hyperref}       
\usepackage{url}            
\usepackage{booktabs}       
\usepackage{amsfonts}       
\usepackage{nicefrac}       
\usepackage{microtype}      
\usepackage{xcolor}         
\usepackage{booktabs}       
\usepackage{graphicx}       
\graphicspath{{../}}
\usepackage[most]{tcolorbox}
\tcbuselibrary{listings, breakable}
\usepackage{booktabs}
\usepackage{multirow}
\usepackage{colortbl}
\usepackage{xcolor}
\usepackage{booktabs}
\usepackage{multirow}
\usepackage{colortbl}
\usepackage{booktabs}       
\usepackage{multirow}       
\usepackage[table,xcdraw]{xcolor}  
\usepackage{amsmath}

\usepackage{amssymb}
\usepackage{mathtools}
\usepackage{amsthm}
\usepackage{algorithm}
\usepackage{algpseudocode}
\usepackage{enumitem}
\usepackage{multirow}
\usepackage{multicol}
\usepackage{graphicx}
\usepackage{tabularx}
\usepackage{makecell}
\theoremstyle{plain}
\usepackage{fancybox}  
\usepackage{fancybox}
\usepackage{xcolor}
\usepackage[most]{tcolorbox}
\usepackage{xcolor}
\usepackage[table]{xcolor}
\usepackage{colortbl}
\usepackage{soul} 
\usepackage{xltabular}
\keepXColumns
\usepackage[table]{xcolor}
\usepackage{colortbl}
\usepackage{soul}
\usepackage{booktabs}
\usepackage{float}
\usepackage[table]{xcolor}
\usepackage{colortbl}
\usepackage{soul} 
\usepackage{array}
\usepackage{placeins}

\theoremstyle{definition}

\theoremstyle{remark}

\newcommand{\hlgreen}[1]{\sethlcolor{naturegreen}\hl{#1}}

\usepackage[preprint]{colm2026_conference}

\usepackage{microtype}
\usepackage{hyperref}
\usepackage{url}
\usepackage{booktabs}

\usepackage{lineno}

\definecolor{darkblue}{rgb}{0, 0, 0.5}
\hypersetup{colorlinks=true, citecolor=darkblue, linkcolor=darkblue, urlcolor=darkblue}

\title{PLUME: Parameter-Efficient Personalization of Large Language Models via Low-Rank User Modulation in Shared Subspaces}

\author{\parbox{0.95\textwidth}{\centering
\textbf{Xinyu Li, Hao Zhou, Jianfeng Zhu, Julina Maharjan, Ruixin Guo, Feodor Dragan, Ruoming Jin} \\
\normalfont
Department of Computer Science \\
Kent State University \\
Kent, OH 44242, USA \\
\texttt{\{xli74,hzhou6,jzhu10,jmaharja,rguo5,fdragan,rjin1\}@kent.edu}
}
}

\begin{document}

\ifcolmsubmission
\linenumbers
\fi

\maketitle
\fancyhead{}

\begin{abstract}
Personalizing large language models (LLMs) is essential for delivering AI assistance that aligns with individual users’ styles, intents, and preferences. While per-user fine-tuning can substantially enhance personalization quality, it introduces significant parameter and storage overhead, limiting scalability to large user populations. We propose \textbf{PLUME} (Personalized Low-Rank Adaptation through User Modulation and Shared Subspace), a lightweight framework that achieves efficient and expressive per-user adaptation by leveraging a shared task-specific subspace.
Specifically, PLUME first learns a global task subspace from aggregated user data. Personalization is then achieved by training only a lightweight small square matrix within this subspace, enabling each user to obtain a tailored model while keeping shared components fixed. Cross-layer shared parameters and rank-1 residual terms are further introduced to significantly reduce redundancy while maintaining expressiveness. Experiments on multiple personalized text generation benchmarks demonstrate that PLUME achieves comparable or superior performance to strong baselines, while reducing per-user parameters by over 95\%. These results establish shared-subspace modulation with minimal residuals as a scalable and semantically grounded approach to LLM personalization. Our code is available at \url{https://github.com/zhouhao0218/PLUME}
\end{abstract}

\input{text/Introduction}
\input{text/Preliminaries}
\input{text/Methods}
\input{text/Experiments}

\input{text/Conclusion}

\bibliography{colm2026_conference}
\bibliographystyle{colm2026_conference}


\clearpage
\newpage
\appendix

\input{text/Appendix_arxiv}



\end{document}

%% file: math_commands.tex
\usepackage{amsmath,amsfonts,bm}

\def\eqref#1{equation~\ref{#1}}

\def\1{\bm{1}}

\DeclareMathAlphabet{\mathsfit}{\encodingdefault}{\sfdefault}{m}{sl}
\SetMathAlphabet{\mathsfit}{bold}{\encodingdefault}{\sfdefault}{bx}{n}



%% file: text/Introduction.tex
\vspace{-6mm}
\section{Introduction}
\vspace{-3mm}
General-purpose large language models (LLMs) have achieved remarkable success across a wide range of natural language tasks. By pre-training on web-scale text corpora, modern LLMs such as GPT, LLaMA, Gemini, and their variants ~\cite{zhao2023survey} acquire strong general-purpose language understanding and generation capabilities, enabling impressive performance spanning QA dialogue, machine translation, and logical reasoning ~\cite{xu2024contrastive, abbasiantaeb2024let, ferrag2025llm}.
However, these models are typically one-size-fits-all, treating all users the same and aiming to generally fit for tasks. This has spurred growing interest in personalized LLM, where an LLM’s outputs are tailored to an individual user’s style, preferences, or context towards an individual's specific domain, task, and environment ~\cite{fan2024survey,zhao2023survey}. Indeed, user-level personalization is increasingly viewed as crucial in applications like web-based QA assistants~\cite{wu2025webwalker}, education ~\cite{chu2025llm}, and healthcare ~\cite{bajwa2021artificial}. Among these diverse applications, personalized text generation represents a critical frontier in LLM research. Users increasingly demand AI systems that reflect their individuality rather than merely produce what they intend to say.

In response to these demands, a substantial wave of research towards personalized LLMs has emerged.
Broadly speaking, existing approaches can be divided into two paradigms.
The first category, \textit{prompt-based approaches} ~\cite{salemi2023lamp, liu2021pre, wang2023learning, kang2023llms, qiu2025measuring}, attempts to achieve personalization through carefully constructed prompts that are built upon retrieved user history, profiles, or interaction logs.
While such designs offer simplicity and interpretability, they heavily rely on explicit and high-quality user signals and are constrained by limited context length.
The second category, \textit{fine-tuning-based approaches}, directly modifies model parameters, hoping to encode personalized information in LLM itself. Majority of studies ~\cite{zhang2024personalized1, zhao2025nextquill} choose to fine-tune a universal model across all user data, hoping to implicitly capture individual preferences deeply mined in different individuals' data via a single run; however, such models tend to blur user distinctions and underperform in fine-grained personalization.

More recently, a small but growing body of work has explored the per-user model paradigm~\cite{tan2024democratizing, bu2025personalized}, in which each individual maintains a dedicated model or adapter.
Despite its advantages in personalization peroformance, this paradigm suffers from storage overhead when scaling to thousands or millions of users. Moreover, given the limited personal writing data, these adapters are at high risk of overfitting. This leads to a central challenge for personalized language modeling in writing assistance: \textit{How to achieve user-level adaptability that captures stylistic nuance and expressiveness without compromising storage efficiency?}

To bridge this gap, we propose \textbf{PLUME} (Personalized Low-rank Adaptation through User Modulation and Shared Subspace), a lightweight yet expressive framework for personalized LLM fine-tuning in writing tasks. 
PLUME first performs \textit{global task adaptation} by training a non-personalized LoRA module on the aggregated data from all users.
The resulting low-rank LoRA matrices capture task-specific knowledge and establish a compact \textit{shared latent subspace} that serves as a global foundation for subsequent personalization.
Building upon this shared representation, PLUME introduces an User-Conditioned Subspace Mixer (USM), which is inserted between the pre-trained LoRA factors to modulate the shared subspace according to each user’s unique preferences.
To further reduce the parameter redundancy, PLUME incorporates a Personalized Cross-Layer Shared (PCLS) module shared across layers, thereby eliminating redundant per-layer parameterization within each personalized model.
Finally, a lightweight rank-1 residual component (Resid) is attached to provide fine-grained layer-specific correction with negligible cost.
Through the principled composition of these components, PLUME achieves highly expressive personalization while significanly reducing per-user trainable parameters. Our key contributions are as follows:

\vspace{-3mm}
\begin{itemize}[leftmargin=*]
    \item \textbf{New Problem formulation} We introduce the problem such that for the personalized text generation task, how to efficiently compress per-user PEFT parameters while maintaining both personalization effectiveness and model expressiveness, which is critical for practical personalization training.
    \item \textbf{PLUME Framework} We show that classic full LoRA is redundant under personalization settings and propose \textsc{PLUME}, a novel and lightweight personalization framework that alleviates the inefficiency of relatively heavy per-user LoRA adapters. PLUME significantly reduces per-user parameter cost while preserving personalization performance and expressive capacity.
    \item \textbf{Extensive Empirical Validation and Analysis}: We conduct comprehensive experiments on five different tasks from LaMP and LongLaMP benchmarks, covering both short and long form content generation. Experiment results show that our framework could robustly reduce the per-user parameter space to less than 5\% of standard LoRA while maintaining or even improving personalization performance.
\end{itemize}

\begin{figure*}[htbp]  
    \centering
    \includegraphics[width=\textwidth]{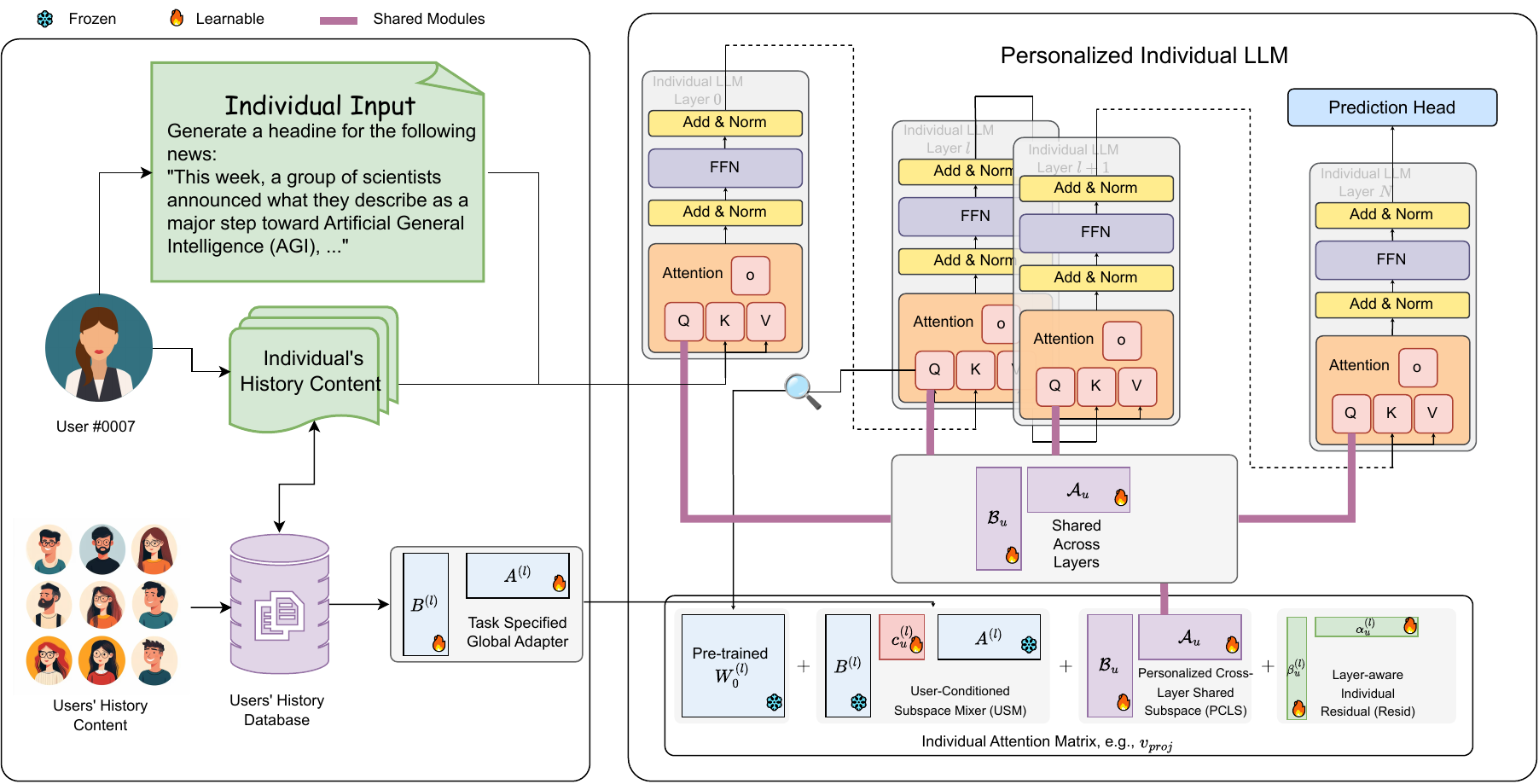}
    \caption{Overall System Architecture. PLUME consists of four components: (i) globally task-specific non-personalized LoRA modules (ii) User-Conditioned Subspace Mixer (USM): $c_u^{(l)}$ between global LoRA factors, (iii) Personalized Cross-Layer Shared Subspace (PCLS): $(\mathcal{A}_u, \mathcal{B}_u)$, and (iv) Layer-aware rank-1 residuals $\alpha_u^{(l)} \beta_u^{(l)}$. Through the principled composition of these complementary modules, PLUME achieves personalized parameter compression while maintaining competitive or superior performance.}
    \vspace{-5mm}
    \label{fig: Overview of PLUME}
\end{figure*}

%% file: text/Preliminaries.tex
\vspace{-5mm}
\section{Problem and Preliminaries}
\vspace{-3mm}
\subsection{Problem Formulation}
\vspace{-4mm}

We consider the task of \textbf{personalized text generation}, 
where the goal is to find a unique adapter for each individual to generate user-specific textual responses conditioned on both the current query and contextual information.

\smallskip
\noindent\textbf{Data Construction.}
Let there be $N$ users denoted by $\{u_1, u_2, \dots, u_N\}$.  
Each user $u_i$ has $M_i$ pairs of history of query–response:
\[
\mathcal{H}_i = \{(q_{i,1}, a_{i,1}), (q_{i,2}, a_{i,2}), \dots, (q_{i,M_i}, a_{i,M_i})\},
\]
where $q_{i,j}$ represents the $j$-th query describing a specific task or instruction, 
and $a_{i,j}$ denotes the corresponding ground-truth response.  
Each response is tokenized as:
\[
a_{i,j} = (a_{i,j}^{(1)}, a_{i,j}^{(2)}, \dots, a_{i,j}^{(T_{i,j})}),
\]
where $a_{i,j}^{(t)}$ denotes the $t$-th token and $T_{i,j}$ its token-wise length.

To both emulate the real deployment regime and conform to the established best practice of few-shot in-context prompting, for each query $q_{i,j}$, we retrieve its top-$k$ similar historical queries $\mathcal{T}_{i,j}
=
\text{TopK}\big(\text{Sim}(q_{i,j}, q_{i,\cdot})\big)
$ from the same user and pair them with their corresponding responses to construct the personalized context:
\begin{equation*}
\mathcal{C}(q_{i,j})
=
\{(q_{i,t}, a_{i,t}) \mid t \in \mathcal{T}_{i,j}\}
\end{equation*}

where $\text{Sim}(\cdot)$ could be any function that measures query-level similarity.  
Each instance is therefore represented as a triplet 
$(q_{i,j}, \mathcal{C}(q_{i,j}), a_{i,j})$.

\smallskip
\noindent\textbf{Training Objective.} We adopt Decoder-Only LLMs for training where a pre-trained large language model parameterized by $\Theta$ serves as the shared backbone, 
and each user $u_i$ is associated with a lightweight personalized adapter parametrized by $\theta_i$.  
The full model for user $u_i$ is denoted as $f_{\Theta, \theta_i}$, 
which conditions generation on both the shared backbone and user-specific adaptation.

For user $u_i$, the model is trained to predict each token in the user’s response given all previous tokens, the current query, and the retrieved context:
\begin{equation}\label{eq: train_ojb}
\begin{split}
&\mathcal{L}_{\text{train}}(\Theta, \theta_i)
= - \sum_{j=1}^{M_i} 
\sum_{t=1}^{T_{i,j}} 
\\ \log  &P_{f_{\Theta, \theta_i}}\big(
a_{i,j}^{(t)} \mid a_{i,j}^{(<t)}, q_{i,j}, \mathcal{C}(q_{i,j}), \mathcal{A}_{\text{other}}
\big),
\end{split} 
\end{equation}

where $\mathcal{A}_{\text{other}}$  represents auxiliary information from other users (Note: We formulate a general LLM personalization task in Eq.\ref{eq: train_ojb}, where $\mathcal{A}_{\text{other}}$ could be in any form, such as other user’s textual histories. In this work, we encode $\mathcal{A}_{\text{other}}$ into global Space $A^{(l)}$ and $B^{(l)}$ as a learned task-specific representation shared by all users).  
During optimization, $\Theta$ is shared or fixed, while $\theta_i$ is optimized to minimize the user-specific loss:
\begin{equation}\label{eq: theta star}
\theta_i^* = 
\arg\min_{\theta_i} \mathcal{L}_{\text{train}}(\Theta, \theta_i), 
\quad i = 1, \dots, N.
\end{equation}

\vspace{-3mm}
\smallskip
\noindent\textbf{Inference.} 
At test time, given a new query $q_i^{\text{test}}$ for user $u_i$, 
we retrieve its top-$k$ similar query–response pairs from $\mathcal{H}_i$ 
to form $\mathcal{C}(q_i^{\text{test}})$.  
The shared backbone combined with the learned personalized parameters 
is then used to generate the predicted response:
\[
\hat{a}_i^{\text{test}} =
f_{\Theta, \theta_i^*}\big(q_i^{\text{test}}, \mathcal{C}(q_i^{\text{test}}), \mathcal{A}_{\text{other}}\big),
\]
and the output $\hat{a}_i^{\text{test}}$ is compared with the ground-truth $a_i^{\text{test}}$ using standard evaluation metrics.

\begin{table}[t]
\centering
\small
\vspace{-3mm}
\renewcommand{\arraystretch}{0.9}
\setlength{\tabcolsep}{4pt}
\begin{tabular}{p{0.18\linewidth} p{0.72\linewidth}}
\toprule
\textbf{Symbol} & \textbf{Description} \\
\midrule
$i$ & User index, $i = 1, \dots, N$. \\
$j$ & Query–response pair index for user $u_i$. \\
$t$ & Token index within a response sequence. \\
$q_{i,j}$ & The $j$-th query of user $u_i$. \\
$a_{i,j}$ & Ground-truth response corresponding to $q_{i,j}$. \\
$\mathcal{H}_i$ & Set of all historical $(q, a)$ pairs for user $u_i$. \\
$\mathcal{C}(q_{i,j})$ & Top-$k$ similar $(q,a)$ pairs retrieved from $\mathcal{H}_i$. \\
$\mathcal{A}_{\text{other}}$ & Auxiliary information (optionally) derived from other users’ histories. \\
$\Theta$ & Parameters of the shared base language model. \\
$\theta_i$ & Personalized parameters (adapter) for user $u_i$. \\
$f_{\Theta, \theta_i}$ & Personalized LLM combining shared backbone and user's PEFT module. \\
$(\cdot)_u$ & Subscript indicating \emph{user-specific} parameters. \\
$(\cdot)^{(l)}$ & Superscript indicating the \emph{$l$-th layer} in the model. \\
\bottomrule
\end{tabular}
\caption{Notation Summary.}
\vspace{-3mm}
\label{tab:notation}
\end{table}


\vspace{-3mm}
\subsection{One-PEFT-per-User Personalization Paradigm}

\begin{figure}[t]
    \centering
    \begin{subfigure}[t]{0.49\linewidth}
        \centering
        \raisebox{2.5mm}{
            \includegraphics[width=\linewidth]{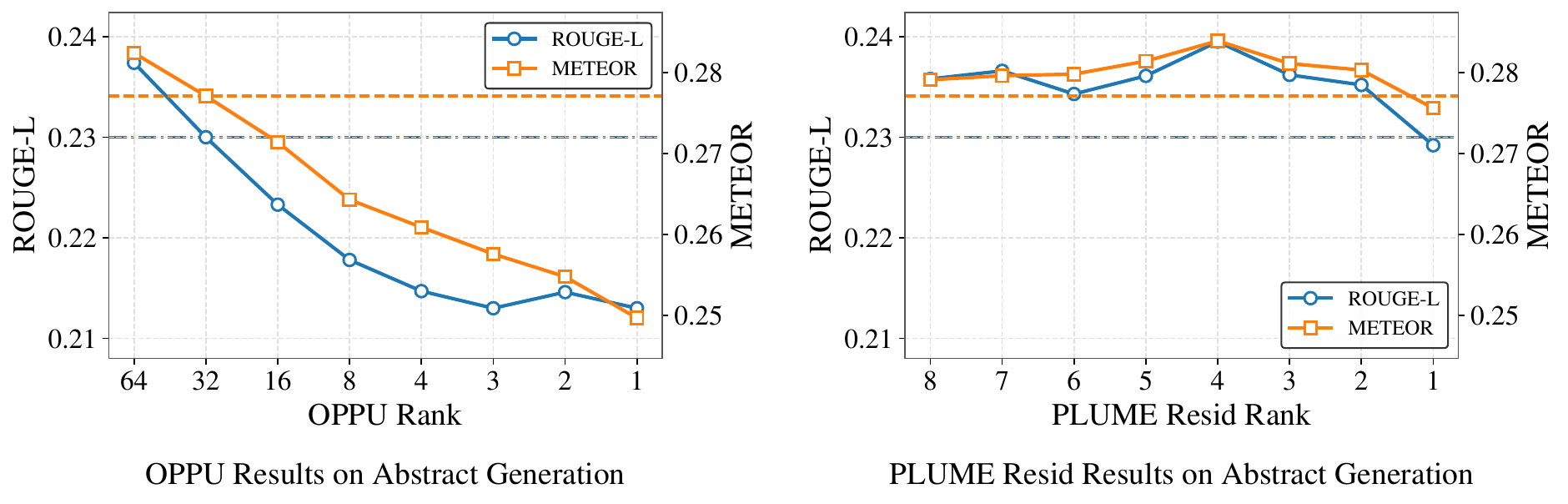}
        }
        \caption{}
        \label{fig:OPPU_performance}
    \end{subfigure}
    \hfill
    \begin{subfigure}[t]{0.49\linewidth}
        \centering
        \includegraphics[width=\linewidth]{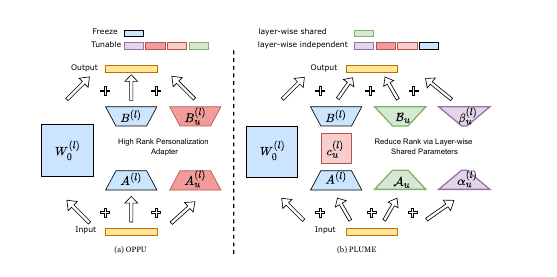}
        \caption{}
        \label{fig:OPPU_vs_PLUME}
    \end{subfigure}
    \caption{\textbf{(a)} OPPU requires a high rank to maintain performance. \textbf{(b)} PLUME achieves comparable results even at low ranks by using layer-wise shared parameters and lightweight user-specific modules.}
    \vspace{-5mm}
\end{figure}

We follow the \textit{One PEFT per User}     (OPPU~\cite{tan2024democratizing}) training framework, 
which assigns an independent PEFT module to each user for personalized fine-tuning. 

Concretely, OPPU first ignores user differences and trains a task-specific global LoRA adapter 
using the union of all users’ training data 
$\bigcup_i \mathcal{H}_i$, resulting in a shared adapter $\Delta\Theta$.  
This stage corresponds to optimizing
\vspace{-3mm}
\begin{equation}
\Delta\Theta^* = 
\arg\min_{\Delta\Theta}
\sum_{i=1}^{N} 
\mathcal{L}_{\text{train}}(\Theta, \Delta\Theta; \mathcal{H}_i),    
\end{equation}
\vspace{-3mm}

The Second step simply follows Eq ~\ref{eq: theta star},  
but with merged model $\Theta\leftarrow\Theta + \Delta\Theta^*$.  

If we further examine this paradigm through the lens of a specific target module, e.g., $q_{\text{proj}}$,  
the resulting parameter composition becomes more explicit. Typically, the effective weight matrix on a layer can be expressed as:
\begin{equation}
    W_{u}^{(l)} = W_{0}^{(l)} + s\, B^{(l)} A^{(l)} + s^{\prime}\, B_{u}^{(l)} A_{u}^{(l)},
\end{equation}
where $W_{0}^{(l)}$ denotes the shared base model weight,  
$(A^{(l)}, B^{(l)})$ represents the global LoRA adapter trained from the aggregated non-personalized data $\bigcup_i \mathcal{H}_i$,  
and $(A^{(l)}_{u}, B^{(l)}_{u})$ denotes the user-specific LoRA adapter paramterized from $\theta_i$.

Vanilla OPPU provides a straightforward way to incorporate user-specific knowledge, nevertheless, it faces two major limitations.  
First, maintaining a separate LoRA module for every user leads to prohibitively high storage and memory costs, especially when the number of users is large.  
Second, since each user’s training data can vary drastically in scale and quality, independently optimizing a full-rank LoRA for every user often causes overfitting or unstable personalization performance.  
As illustrated in Figure \ref{fig:OPPU_performance}, the performance of OPPU drops significantly when the individual LoRA rank is reduced.

These issues highlight the need for a more parameter-efficient and stable approach to capture user-specific preferences without requiring a full LoRA module per user.

\subsection{Notation Summary}
Table~\ref{tab:notation} summarizes the key symbols used in this paper, 
covering both the preceding problem formulation and the following method sections.

%% file: text/Methods.tex
\vspace{-3mm}
\section{Method}
\label{sec: method}

In this section, we propose \textbf{PLUME}, 
a novel framework designed to reduce redundant personalized parameters while preserving personalization expressivity.  
Figure~\ref{fig: Overview of PLUME} illustrates the overall architecture of PLUME. We will start from our key observation, comparison with vanilla OPPU and progressively builds our efficient yet effective model.
\vspace{-3mm}
\subsection{PLUME Framework}
\noindent\textbf{Revisiting Task Space.}
From a linear algebraic perspective, the task-specific LoRA update on layer $l$ can be expressed as 
$\Delta W^{(l)} = s\, B^{(l)} A^{(l)}$, where $A^{(l)}$ projects inputs into a low-dimensional \emph{task subspace}, 
and $B^{(l)}$ expands the projected representation back to the model’s hidden space. Hence, for any input $x$, the LoRA output $\Delta W^{(l)}x = s\, B^{(l)}A^{(l)}x$ always resides in the column space $\mathrm{col}(B^{(l)})$. 
OPPU extends this by assigning each user an individual adapter $(A_u^{(l)}, B_u^{(l)})$, which effectively introduces a new output subspace basis $\mathrm{col}(B^{(l)}_u)$ specific to each user.   
From this view, OPPU increases the expressive capacity of the model by expanding the dimensional coverage of $\mathrm{col}(B^{(l)})$ across users, 
thereby enhancing personalization.  

However, this completely \enquote{\emph{separates}}
the shared task space $\mathrm{col}(B^{(l)})$ and the augmented individual space $\mathrm{col}(B^{(l)}_u)$, leaving the shared adapter only learns how to project tokens into the task subspace, 
while not being able to control how different users \emph{behave} within the collaboratively trained \emph{existing shared task subspace} $\mathrm{col}(B^{(l)})$. This motivates us to allow user-dependent modulation of the shared subspace.

\smallskip
\noindent\textbf{User-Conditioned Subspace Mixer ($c_u$).}
Intuitively, since users may differ in style, intent, or preference, 
the way they combine or emphasize task-subspace directions should vary. To capture user-specific task space utilization, we insert a lightweight matrix $c_u^{(l)} \in \mathbb{R}^{r_g \times r_g}$ between $A^{(l)}$ and $B^{(l)}$:
\begin{equation}
\Delta W_u^{(l)} = s\, B^{(l)} c_u^{(l)} A^{(l)}.
\label{eq:usm}
\end{equation}
We term $c_u^{(l)}$ a \emph{User-Conditioned Subspace Mixer (USM)}.  
It can be interpreted as a re-indexing or re-weighting operator that re-combines latent task directions inside the shared LoRA subspace.  
Thus, $c_u^{(l)}$ enables personalized modulation of how a user exploits the common task subspace $\mathrm{col}(B^{(l)})$,
offering additional expressivity with negligible parameter cost.


\smallskip
\noindent\textbf{PLUME.}
Before we move on to a more compact model, we first assess how much expressivity the USM alone provides, we introduce a lightweight per-layer residual term aiming to approximate the representational power of high-rank OPPU:
\begin{equation}
W_u^{(l)} = W_0^{(l)} + s\, B^{(l)} c_u^{(l)} A^{(l)} + s'\, b_u^{(l)} a_u^{(l)}.
\label{eq:plume_wo_shared}
\end{equation}
Here $(a_u^{(l)}, b_u^{(l)})$ follow the same formulation as $(A_u^{(l)}, B_u^{(l)})$, but with a reduced individual rank $r_{\text{ind}}$.  
Empirically, we find that when $r_{\text{ind}}\!\approx\!4$, PLUME already achieves performance comparable to OPPU with rank~64, as shown in Fig~\ref{fig:OPPU_performance}.  
Interestingly, further increasing $r_{\text{ind}}$ provides no additional gains—indicating that once the shared task subspace is efficiently utilized via $c_u^{(l)}$, a large individual subspace becomes redundant and inefficient, leading to overfitting.  
This observation suggests that there may still exist parameter redundancy even within the reduced adapters.  
Thus, we further push PLUME to the extreme—seeking the most \emph{compact yet expressive} form of individual representation. As illustrated in Fig \ref{fig:OPPU_vs_PLUME}, our proposed novel shared subspace mechanism enables PLUME to reduce the individual parameter while maintaining highly compatible performance.

\begin{figure}
    \centering
    \includegraphics[width=0.6\linewidth]{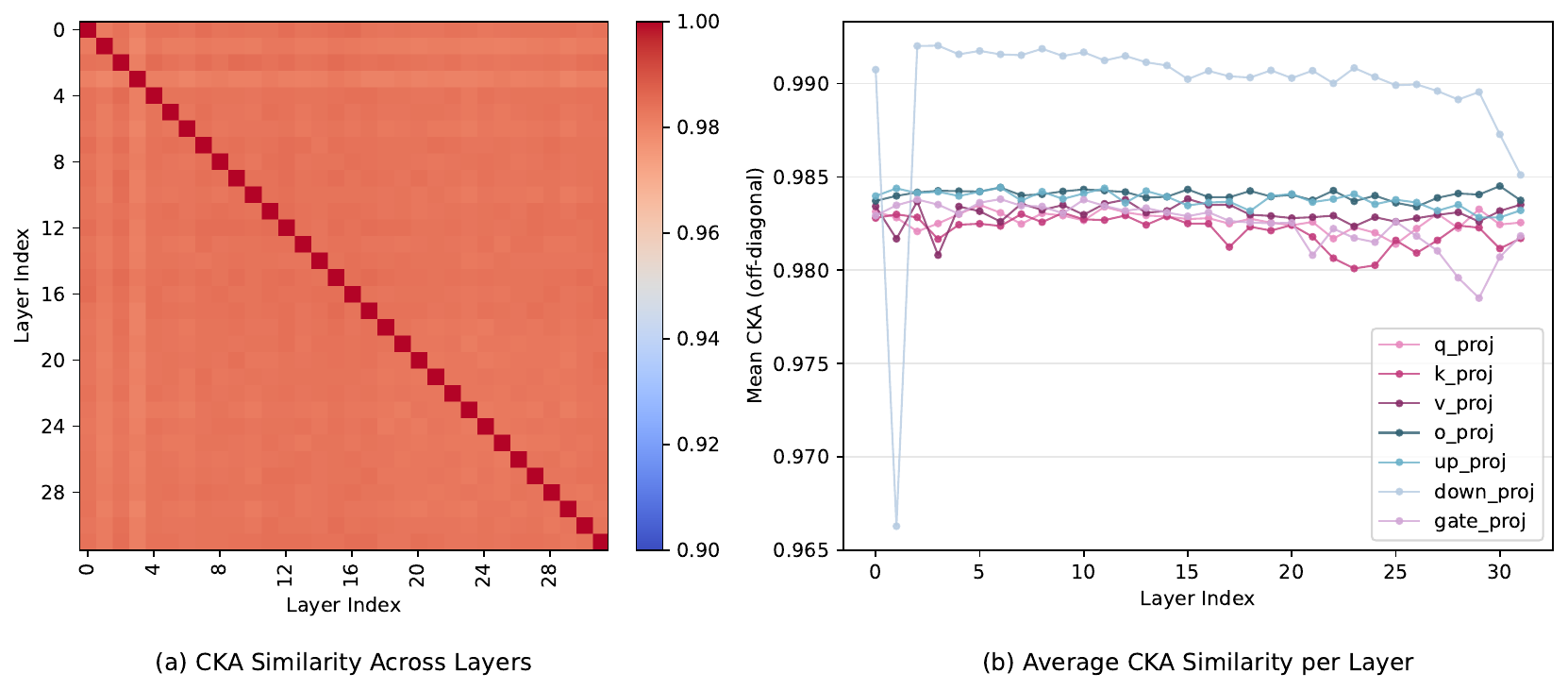}
   \caption{CKA analysis of a user's LoRA parameters. 
(a) Pairwise CKA across layers; (b) Average CKA similarity per layer. 
Strong inter-layer similarity reveals significant redundancy, supporting PLUME’s shared-layer design.}
\vspace{-3mm}
    \label{fig:cka}
\end{figure}

\subsection{Parameter Reduction via Shared Subspace} 
\smallskip
\noindent\textbf{CKA Analysis for Layer-Wise Representation Similarity for Personalized Model}
Inspired by prior work~\cite{he2025rasa, kopiczko2023vera, zhoubslora} that sharing weights across layers can enhance model expressiveness while reducing parameter usage, we investigate whether different layers in personalized LoRA models actually learn similar user-specific representations. To this end, we perform a \emph{Centered Kernel Alignment} (CKA), 
a widely used approach for comparing neural representations across layers and models ~\cite{kornblith2019similarity, liu2025spectral} to quantify layer-wise representational similarity across the personalized adapters (See Appendix~\ref{sec: cka_analysis} for more about CKA).


As shown in Figure~\ref{fig:cka}, OPPU-trained adapters exhibit remarkably high inter-layer CKA values (often exceeding $0.9$), 
indicating that the personalized LoRA subspaces learned by different layers are strongly aligned.  
This suggests that the layer-wise residuals $\{a_u^{(l)}, b_u^{(l)}\}$ tend to encode similar directions of user-specific variation, 
leading to redundant parameterization across depth.  
Motivated by this finding, we hypothesize that a shared subspace across layers could capture the dominant personalized factors more efficiently.

\smallskip
\noindent\textbf{PLUME-s.}
Building on this insight, we observe that the per-layer residuals $(a_u^{(l)}, b_u^{(l)})$ 
often act similarly to layer-specific biases that adjust the shared projection directions.  
Accordingly, we decompose $(a_u^{(l)}, b_u^{(l)})$ into two functional components: 
(a) a \emph{Personalized Cross-Layer Shared Subspace (PCLS), $(\mathcal{A}_u, \mathcal{B}_u)$}) capturing cross-layer user-specific residuals in a low-rank shared form, and 
(b) \emph{a rank-1 layer-wise Residual (Resid)}, $(\alpha_u^{(l)}, \beta_u^{(l)})$ that enables fine-grained local adaptation.
The resulting formulation becomes:
\begin{equation}
W_u^{(l)} = W_0^{(l)} 
+ s\, B^{(l)} c_u^{(l)} A^{(l)} 
+ s'\, \mathcal{B}_u \mathcal{A}_u
+ \beta_u^{(l)} \alpha_u^{(l)}.
\label{eq:plume_w_shared}
\end{equation}
Here $(\mathcal{A}_u,\mathcal{B}_u)$ with rank $r_{\textbf{sh}}$ parameterize the Layer-Shared Personalized Subspace reused across all layers, 
while $(\alpha_u^{(l)},\beta_u^{(l)})$ serve as the rank-1 layer refinements for fine adjustment. 
Extensive experiments demonstrate that leveraging a shared residual subspace allows substantial reduction of individual residual ranks without compromising performance, and can even outperform OPPU in certain settings (see Sec.~\ref{sec:experiments}).

\smallskip
\noindent\textbf{Summary and Parameter Efficiency.}
Let $L$ be the number of layers and $r_g$ the global LoRA rank, and $d$ the embedding vector dimension. In conclusion, PLUME adaptively unifies four complementary components: a global task-specific adapter $(A^{(l)}, B^{(l)})$ shared by all users,  
and three lightweight personalized modules.  
(1) The User-Conditioned Subspace Mixer (USM) $c_u^{(l)}$ adaptively reuses the global task space with $O(L r_g^2)$ parameters.  
(2) The Personalized Cross-Layer-Shared Personalized Subspace (PCLS) $(\mathcal{A}_u,\mathcal{B}_u)$ captures cross-layer personalization compactly with $O(r_{\text{sh}} d)$ parameters.  
(3) The rank-1 personalized residuals $(\alpha_u^{(l)},\beta_u^{(l)})$ restore fine local flexibility using $O(L d)$ parameters. In total, the per-user complexity is 
\(
O(L r_g^2 + r_{\text{sh}} d + L d),
\)
compared to $O(L d r_{\text{oppu}})$ for OPPU’s layer-wise independent adapters.  When $r_{\text{sh}} \!\ll\! r_{\text{oppu}}$, the parameter reduction becomes particularly significant, see Table~\ref{tab:param_efficiency} for practical analysis. 

%% file: text/Experiments.tex
\vspace{-3mm}
\section{Experiments}
\label{sec:experiments}
\vspace{-3mm}
\begin{table*}[t]
\centering
\vspace{-3mm}
\resizebox{\textwidth}{!}{%
\begin{tabular}{llccccccccccccccc}
\toprule
\multirow{3}{*}{Type} & \multirow{3}{*}{Method} 
& \multicolumn{9}{c}{\textbf{Long Content Generation}} 
& \multicolumn{6}{c}{\textbf{Short Content Generation}} \\
\cmidrule(lr){3-11} \cmidrule(lr){12-17}
& & \multicolumn{3}{c}{Abstract Generation} 
& \multicolumn{3}{c}{Product Review} 
& \multicolumn{3}{c}{Topic Writing} 
& \multicolumn{3}{c}{News Headline} 
& \multicolumn{3}{c}{Scholarly Title} \\
\cmidrule(lr){3-5} \cmidrule(lr){6-8} \cmidrule(lr){9-11} 
\cmidrule(lr){12-14} \cmidrule(lr){15-17}
& & R-1 & R-L & MTR & R-1 & R-L & MTR & R-1 & R-L & MTR & R-1 & R-L & MTR & R-1 & R-L & MTR \\
\midrule
\multirow{6}{*}{\textbf{Non-Personalized}}
& BASE & 0.3497 & 0.1688 & 0.2408 & 0.3397 & 0.1389 & 0.2335 & 0.2892 & 0.1238 & 0.2050 & 0.1313 & 0.1158 & 0.0962 & 0.3905 & 0.3165 & 0.4084 \\
& LoRA & 0.3491 & 0.2036 & 0.2456 & 0.3877 & 0.2323 & 0.2707 & 0.2655 & 0.1378 & 0.1732 & 0.2188 & 0.2010 & 0.1939 & 0.4648 & 0.4138 & 0.4228 \\
& PiSSA & 0.3523 & 0.1995 & 0.2536 & 0.3964 & 0.2326 & 0.2826 & 0.2854 & 0.1393 & 0.1927 & 0.2230 & 0.2056 & 0.1999 & 0.4801 & 0.4266 & 0.4351 \\
& AdaLoRA & 0.3503 & 0.2089 & 0.2395 & 0.3419 & 0.2056 & 0.2323 & 0.2279 & 0.1212 & 0.1530 & 0.2299 & 0.2113 & 0.1892 & 0.4621 & 0.4203 & 0.3940 \\
& QLoRA & 0.3450 & 0.2001 & 0.2394 & 0.3940 & 0.2339 & 0.2769 & 0.2697 & 0.1399 & 0.1762 & 0.2224 & 0.2037 & 0.1958 & 0.4628 & 0.4101 & 0.4196 \\
& LoRA-One & 0.3780 & 0.2317 & 0.2750 & 0.3800 & 0.2320 & 0.2680 & 0.2930 & 0.1410 & 0.2110 & -- & -- & -- & 0.4480 & 0.4030 & 0.4070 \\
\midrule
\multirow{5}{*}{\textbf{Personalized}}
& RAG & 0.3512 & 0.1743 & 0.2567 & 0.3389 & 0.1454 & 0.2336 & 0.2936 & 0.1264 & 0.2199 & 0.1470 & 0.1304 & 0.1047 & 0.3982 & 0.3229 & 0.4050 \\
& OPPU & 0.4147 & 0.2374 & 0.2824 & 0.4416 & \textbf{0.2481} & \textbf{0.3165} & 0.3191 & 0.1510 & 0.2126 & 0.2383 & 0.2178 & 0.2008 & 0.5146 & 0.4510 & \textbf{0.4365} \\
& CoPE & 0.3779 & 0.2247 & 0.2530 & 0.3600 & 0.2385 & 0.2645 & 0.2303 & 0.1386 & 0.1754 & 0.2324 & 0.2101 & 0.1963 & 0.4741 & 0.4076 & 0.4133 \\
& \textbf{PLUME} & \textbf{0.4168} & \textbf{0.2395} & \textbf{0.2839} & \textbf{0.4423} & 0.2460 & 0.3161 & \textbf{0.3217} & \textbf{0.1534} & \textbf{0.2172} & 0.2150 & 0.1974 & 0.1778 & 0.5075 & 0.4459 & 0.4246 \\
& \textbf{PLUME-s} & 0.4117 & 0.2322 & 0.2803 & 0.4369 & 0.2434 & 0.3107 & 0.3176 & 0.1517 & 0.2131 & \textbf{0.2427} & \textbf{0.2219} & \textbf{0.2039} & \textbf{0.5177} & \textbf{0.4537} & 0.4359 \\
\bottomrule
\end{tabular}
}
\caption{Performance comparison on Mistral-7B across five personalized text generation tasks. Bold numbers indicate the best results within each task. \emph{PLUME} refers to Eq~\ref{eq:plume_wo_shared}, and \emph{PLUME-s} refers to Eq~\ref{eq:plume_w_shared}. Results of LoRA-One on the News Headline task are abnormally low and thus excluded from reporting.}
\label{tab:main_results}
\end{table*}

\begin{table*}[t]
\small
\centering
\vspace{-2mm}
\begin{tabular}{lcccc}
\toprule
\textbf{Method} 
& \textbf{\#Params} 
& \textbf{\% of OPPU}
& \textbf{Memory (GB)}
& \textbf{Runtime} \\
\midrule
OPPU     & 167{,}772{,}160 & 100\% & 0.6250 & 4h53m \\
CoPE     & 167{,}772{,}160 & 100\% & 0.1016 & 4h56m \\
PLUME    & 11{,}403{,}264  & 6.80\% & 0.0425 & 4h30m \\
PLUME-s  & 3{,}702{,}784   & 3.11\% & 0.0156 & 4h42m \\
\bottomrule
\end{tabular}
\caption{Per-user parameter cost and running time comparison on Mistral-7B backbone. PLUME achieves substantial parameter savings while maintaining personalization capacity.}
\vspace{-6mm}
\label{tab:param_efficiency}
\end{table*}

In this section, we conduct experiments to systematically investigate the following research questions:\\
RQ1: How does PLUME perform against non-personalized and personalized PEFT benchmarks in various personalized text generation tasks?\\ 
RQ2: How much efficiency gain does PLUME achieve?\\
RQ3: How do the different components in PLUME affect its performance effectiveness?\\
RQ4: How does PLUME behave under varying degrees of parameter reduction, and what is the resulting trade-off between model compactness and personalization performance?

\subsection{Experimental Setup.}
\noindent\textbf{Datasets}
We adopt the widely used benchmarks LongLaMP~\cite{kumar2024longlamp} and LaMP~\cite{salemi2023lamp} to evaluate our model’s ability across both short-form and long-form content personalized content generation tasks. Specifically, we select 3 datasets: Abstract Generation, Product Review, and Topic Writing from LongLaMP, and 2 datasets: News Headline generation, Scholarly Title generation from LaMP. We leave details on data preprocessing and data statistics to Appendix \ref{sec: more data}

\smallskip
\noindent\textbf{Baselines.} 
We compare our methods against a series of strong parameter-efficient fine-tuning baselines under both non-personalized training and personalized training settings. 
For the \textit{non-personalized} setting, we compare against the \textit{base model}, and  
\textit{LoRA}~\cite{hu2021lora}, 
\textit{AdaLoRA}~\cite{zhang2023adalora}, 
\textit{PiSSA}~\cite{meng2024pissa}, 
and \textit{QLoRA}~\cite{dettmers2023qlora} and \textit{LoRA-One}~\cite{zhang2025loraone}
which all adapt model parameters through a shared low-rank adapter across users. 
For the \textit{personalized} setting, we consider retrival-based approach such as \textit{RAG}~\cite{tan2024democratizing}, training-based baseline 
\textit{OPPU}~\cite{tan2024democratizing}, 
and a more recent work \textit{CoPE}~\cite{bu2025personalized}.  
Finally, we evaluate our two variants: \textit{PLUME} and \textit{PLUME-s}.

\noindent\textbf{Implementation}
To ensure fairness, all models are evaluated with a consistent LoRA configuration (rank = 64) across both non-personalized and personalized settings, including the OPPU and CoPE baselines. To demonstrate the generalizability and robustness of our approach, we evaluate PLUME on two distinct open-source large language models: LLaMA2-7B~\cite{touvron2023llama2} and Mistral-7B-Instruct-v0.2~\cite{jiang2023mistral}.  As illustrated in Figure~\ref {fig:Ablation_Resid_Rank} in the later discussion, \textit{PLUME} already reaches competitive results with a very small rank, so we report \textit{PLUME} results with residual rank $4$, and \textit{PLUME-s} with residual rank $1$ and shared rank as low as $8$.
Due to space constraints, we present the results using the Mistral-7B backbone in the main text and leave more experiment results and details in Appendix~\ref{sec: more exp}.

\noindent\textbf{Evaluation Metrics}
In accordance with established practices in prior work~\cite{tan2024democratizing, kumar2024longlamp}, we employ a standard set of automatic metrics ROUGE-1, ROUGE-L~\cite{lin2004rouge}, and METEOR~\cite{banerjee2005meteor} to quantitatively assess the lexical overlap, fluency, and semantic correspondence between generated outputs and reference responses (See more details in Appendix~\ref{sec:appendix_metrics})

\subsection{Main Results}
\noindent\textbf{Overall Perfomance (RQ1)}
Table~\ref{tab:main_results} reports the overall performance across five generation tasks. Overall, personalized methods outperform non-personalized ones across all metrics, confirming the advantage of user-specific adaptation. Within the personalized group, despite using far fewer parameters than strong baselines such as OPPU and CoPE, the PLUME series delivers comparable or better results than other baseline methods. For instance, PLUME attains the highest ROUGE-L on Abstract Generation (0.2395) and comparable performance on other long-text tasks, demonstrating that the personalized module USM would substantially enhances expressiveness. When incorporating the shared subspace, PLUME-s performance slightly improves, e.g., ROUGE-1 rises from 0.2150 to 0.2427 on News Headline Generation Tasks. This indicates that the shared subspace not only compresses parameters but also mitigates redundancy and overfitting issues observed in OPPU.

In summary, PLUME achieves a favorable balance between expressiveness and efficiency. It retains nearly identical generation quality to strong personalized baselines OPPU and CoPE while using only a fraction of their parameters, and even surpasses them on several short-text tasks.


\smallskip
\noindent\textbf{Efficiency Comparison (RQ2)}
One of our main contributions is that PLUM substantially improves parameter efficiency for LLM personalization adapters. Here, we quantify the extent to which this compression frontier can be pushed under practical  setting.
As in Table~\ref{tab:param_efficiency}, compared to OPPU, which requires 168M parameters per user, \textit{PLUME} and \textit{PLUME-s} reduce the cost to only 6.8\% and 3.1\% respectively, achieving over $15\times$–$30\times$ compression while preserving personalization quality and requiring less training time.

\smallskip
\noindent\textbf{Ablation Study: Effectiveness of Components (RQ3)}
\begin{figure}[t]
    \centering
    \begin{subfigure}[t]{0.32\linewidth}
        \centering
        \includegraphics[width=\linewidth, height=6.0cm]{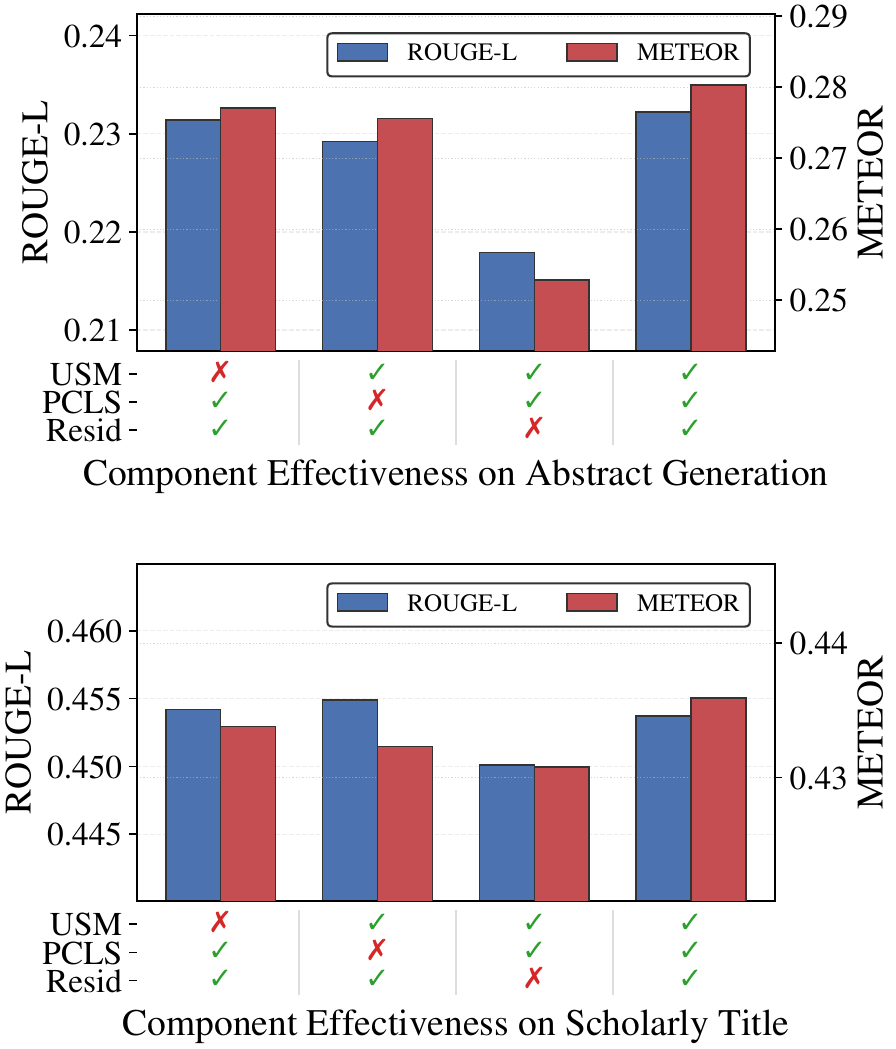}
        \caption{}
        \label{fig:Ablation_Module}
    \end{subfigure}
    \hfill
    \begin{subfigure}[t]{0.32\linewidth}
        \centering
        \includegraphics[width=\linewidth, height=6.0cm]{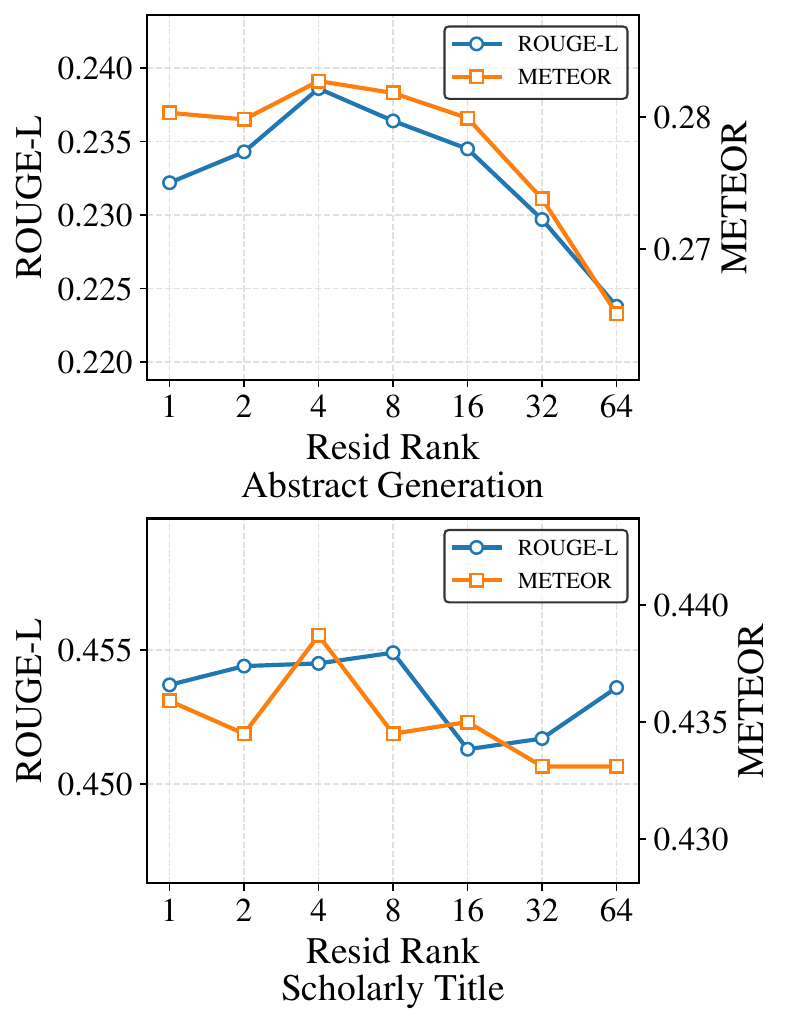}
        \caption{}
        \label{fig:Ablation_Resid_Rank}
    \end{subfigure}
    \hfill
    \begin{subfigure}[t]{0.32\linewidth}
        \centering
        \includegraphics[width=\linewidth, height=6.0cm]{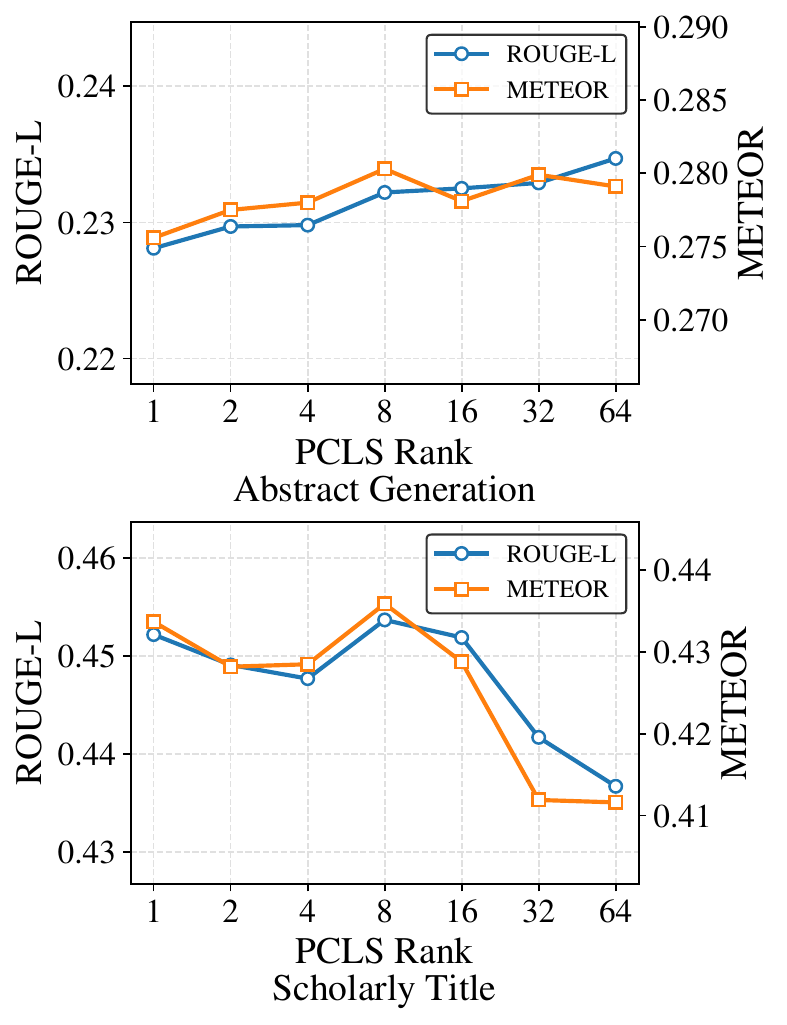}
        \caption{}
        \label{fig:Ablation_PCLS_Rank}
    \end{subfigure}
    \caption{
    \textbf{(a)} Component effectiveness.
    \textbf{(b)} Sensitivity on Resid Rank.
    \textbf{(c)} Sensitivity on PCLS Rank.
    }
    \vspace{-6mm}
    \label{fig:Ablation_All}
\end{figure}
Figure~\ref{fig:Ablation_Module} compares the effects of three core modules: USM, PCLS, and Resid on both long and short text generation tasks.

Across both tasks, removing any single module consistently leads to a drop in ROUGE-L and METEOR, suggesting that all three work jointly to enhance expressiveness, coherence, and efficiency. This consistent trend highlights that the modules are complementary rather than redundant. The relative impact of each module remains consistent across both tasks (Resid > PCLS > USM). Moreover, the effect of the Resid module is more pronounced in long-form generation tasks. Notably, even though Resid employs only a rank-1 update, it still contributes sufficiently to representational richness, showing that lightweight residual paths can complement USM’s user-specific modulation effectively. These results demonstrate that each component is indispensable for achieving a strong balance between personalization, coherence, and parameter efficiency.



\smallskip
\noindent\textbf{Parameter Efficiency VS Model Performance (RQ4)}\\
\noindent\textit{Sensitivity of residual rank.}
To investigate the sensitivity of residual dimensionality, we vary the rank of the residual module. As shown in Figure~\ref{fig:Ablation_Resid_Rank}, performance on Abstract Generation (left) first increases with larger residual ranks and peaks around rank 4, after which both ROUGE-L and METEOR scores gradually decline. This trend indicates that a small residual rank is sufficient to capture user-specific variations, while larger ranks introduce redundancy and overfitting. In contrast, the Scholarly Title task (right) shows much smaller variance across ranks, with performance already strong at rank 1 and exhibiting a slight downward trend thereafter. This validates our conjecture that short-form generation requires fewer personalized parameters and benefits more from PLUME’s shared-subspace regularization.

\smallskip
\noindent\textit{Sensitivity on PCLS rank.}
To further examine the influence of the shared subspace capacity, we vary the rank of PCLS module. As illustrated in Figure~\ref{fig:Ablation_PCLS_Rank}, the results show an opposite trend between the two settings. For Abstract Generation, performance steadily improves as the shared rank increases, indicating that long-form generation benefits from a richer shared subspace capable of modeling broader contextual dependencies and semantic consistency across layers. In contrast, the Scholarly Title task exhibits an inverted trend—performance peaks at a small shared rank and then declines as the rank grows—suggesting that short-form generation requires only limited shared capacity. These contrasting patterns highlight PLUME’s flexibility that its shared subspace can effectively balance generalization and personalization.

%% file: text/Conclusion.tex
\vspace{-5mm}
\section{Conclusion}
\vspace{-3mm}
This paper investigates how to achieve fine-grained user adaptation with better parameter efficiency in the personalized text generation task. To tackle this, we proposed PLUME, a novel framework for personalized LLMs through low-rank user modulation and shared subspaces. The proposed design comprises of the User-Conditioned Subspace Mixer (USM), the cross-layer shared personalized subspace (PCLS), and the rank-1 residuals (Resid) and demonstrates that rich personalization can be achieved with only a fraction of the parameters required by conventional per-user adapters such as OPPU. Comprehensive experiments across five personalized text-generation benchmarks with two different foundation models as backbone show that PLUME consistently matches or exceeds existing personalized LoRA variants while reducing per-user parameters by over 95\%. 

Overall, PLUME provides a more effective and highly compact user representation for personalized LLMs, and we believe it establishes a new research pathway for advancing LLM personalization.

%% file: text/Appendix_arxiv.tex
\appendix
\label{sec: appendix}

\input{text/Related_work}

\section{Data Processing Detail and Data Statistics}
\label{sec: more data}
In preparing the data, we follow the general setup of prior frameworks such as OPPU~\cite {tan2024democratizing} and CoPE~\cite{bu2025personalized} and select 200 users with sufficient interaction histories as our evaluation cohort. For each user, we aggregate all historical interactions and then partition them into training, validation, and test subsets with an 8:1:1 ratio based on temporal ordering. When a personalization prompt requires leveraging user history as examples, we restrict retrieval to the training portion only, ensuring a realistic personalization setup without test leakage. In this case, we employ Contriever~\cite{izacard2021unsupervised} to retrieve the top-K most relevant history entries, which are then included as few-shot demonstrations for generating the target content. This design allows us to assess both the model’s raw personalization ability and its robustness to retrieved history length and quality. Dataset statistics and splits are provided in Table~\ref{tab:dataset_stats}.

\begin{table*}[t]
\centering
\vspace{-3mm}
\begin{tabular}{lccccc}
\toprule
\textbf{Dataset} & \textbf{AG} & \textbf{PR} & \textbf{TW} & \textbf{NH} & \textbf{ST} \\
\midrule
\# Questions & 14,065 & 6,389 & 5,138 & 33,072 & 15,202 \\
Avg Q Length & 332.61 & 881.90 & 585.68 & 188.36 & 509.99 \\
Avg Target Length & 181.44 & 416.73 & 324.66 & 14.95 & 14.54 \\
\bottomrule
\end{tabular}
\caption{Dataset statistics for all tasks. AG = Abstract Generation, PR = Product Review, TW = Topic Writing, NH = News Headline, ST = Scholarly Title.}
\label{tab:dataset_stats}
\end{table*}

\begin{table*}[t]
\centering
\vspace{-3mm}
\resizebox{\textwidth}{!}{
\begin{tabular}{llccccccccccccccc}
\toprule
\textbf{Type} & \textbf{Method} 
& \multicolumn{9}{c}{\textbf{Long Content Generation}} 
& \multicolumn{6}{c}{\textbf{Short Content Generation}} \\ 
\cmidrule(lr){3-11} \cmidrule(lr){12-17}
 &  & \multicolumn{3}{c}{Abstract Generation} & \multicolumn{3}{c}{Product Review} & \multicolumn{3}{c}{Topic Writing} 
 & \multicolumn{3}{c}{News Headline} & \multicolumn{3}{c}{Scholarly Title} \\ 
\cmidrule(lr){3-5} \cmidrule(lr){6-8} \cmidrule(lr){9-11} 
\cmidrule(lr){12-14} \cmidrule(lr){15-17}
 &  & R-1 & R-L & MTR & R-1 & R-L & MTR & R-1 & R-L & MTR 
 & R-1 & R-L & MTR & R-1 & R-L & MTR \\
\midrule
\multirow{6}{*}{\textbf{Non-Personalized}}
 & BASE      & 0.2050 & 0.1092 & 0.1286 & 0.3248 & 0.1390 & 0.2011 & 0.2710 & 0.1204 & 0.1727 & 0.1320 & 0.1157 & 0.0831 & 0.4379 & 0.3556 & 0.4192 \\
 & LoRA      & 0.3358 & 0.1983 & 0.2279 & 0.3602 & 0.2204 & 0.2489 & 0.2465 & 0.1331 & 0.1585 & 0.2106 & 0.1948 & 0.1786 & 0.4474 & 0.4004 & 0.3849 \\
 & PiSSA            & 0.3378 & 0.1910 & 0.2374 & 0.3833 & 0.2276 & 0.2687 & 0.2681 & 0.1359 & 0.1768 & 0.2140 & 0.1970 & 0.1886 & 0.4718 & 0.4207 & \textbf{0.4263} \\
 & AdaLoRA          & 0.3403 & 0.1989 & 0.2328 & 0.2713 & 0.1336 & 0.1595 & 0.2135 & 0.1095 & 0.1325 & 0.2127 & 0.1963 & 0.1718 & 0.4445 & 0.4029 & 0.3744 \\
 & QLoRA            & 0.3260 & 0.1934 & 0.2157 & 0.3668 & 0.2238 & 0.2563 & 0.2401 & 0.1291 & 0.1515 & 0.2062 & 0.1907 & 0.1735 & 0.4482 & 0.4030 & 0.3835 \\
 & LoRA-One         & \textbf{0.4080} & \textbf{0.2700} & \textbf{0.3060} & 0.3750 & 0.2180 & 0.2660 & 0.2720 & 0.1340 & 0.1830 & 0.2150 & 0.1980 & 0.1900 & 0.4700 & 0.4190 & 0.4230 \\
\midrule
\multirow{5}{*}{\textbf{Personalized}}
 & RAG       & 0.3549 & 0.1808 & 0.2435 & 0.3268 & 0.1399 & 0.2111 & 0.2435 & 0.1124 & 0.1607 & 0.1322 & 0.1157 & 0.0941 & 0.4231 & 0.3461 & 0.3881 \\
 & OPPU      & 0.3933 & 0.2214 & 0.2614 & 0.4156 & 0.2369 & 0.2872 & 0.2878 & 0.1428 & 0.1827 & 0.2285 & 0.2091 & 0.1900 & 0.4955 & 0.4364 & 0.4050 \\
 & CoPE             & 0.3320 & 0.2071 & 0.2731 & 0.3263 & 0.2125 & 0.2644 & 0.1607 & 0.1041 & 0.1370 & 0.1779 & 0.1589 & 0.1928 & 0.4030 & 0.3485 & 0.3920 \\
 & \textbf{PLUME} & 0.3962 & 0.2225 & 0.2664 & \textbf{0.4209} & 0.2393 & \textbf{0.2947} & \textbf{0.2944} & \textbf{0.1439} & \textbf{0.1872} & 0.2291 & 0.2100 & 0.1916 & 0.5103 & 0.4487 & 0.4231 \\
 & \textbf{PLUME-s}  & 0.3865 & 0.2169 & 0.2543 & 0.4142 & \textbf{0.2397} & 0.2890 & 0.2790 & 0.1406 & 0.1742 & \textbf{0.2336} & \textbf{0.2155} & \textbf{0.1940} & \textbf{0.5107} & \textbf{0.4512} & 0.4243 \\
\bottomrule
\end{tabular}}
\caption{Performance comparison on LLaMA-2-7B across five personalized text generation tasks. Bold numbers indicate the best
results within each task.}
\label{tab:llama2_results}
\end{table*}

\begin{table*}[t]
\centering
\vspace{-3mm}
\begin{tabular}{lccc}
\toprule
\textbf{\# Target Token} & \textbf{OPPU} & \textbf{PLUME} & \textbf{PLUME-s} \\
\midrule
Bin 1: 79--125   & 120 & 119 (\,-0.83\%\,) & 123 (\,+0.2.5\%\,) \\
Bin 2: 126--140  & 133 & 130 (\,-2.20\%\,) & 135 (\,+1.50\%\,) \\
Bin 3: 141--155  & 140 & 139 (\,-0.71\%\,) & 141 (\,+0.71\%\,) \\
Bin 4: 156--170  & 151 & 150 (\,-0.67\%\,) & 155 (\,+2.64\%\,) \\
Bin 5: 171--245  & 156 & 158 (\,+1.28\%\,) & 166 (\,+6.41\%\,) \\
\bottomrule
\end{tabular}
\caption{Average generated tokens by configuration across ground-truth length bins. Percentages for PLUME and PLUME-s indicate change relative to OPPU.}
\label{tab:gen_len_bins}
\end{table*}

\section{More Experiment}
\label{sec: more exp}

\subsection{Details of Evaluation Metrics}
\label{sec:appendix_metrics}

To improve clarity with standard text generation benchmarks, we provide formal definitions of the automatic evaluation metrics used in our experiments. Let $G$ denote the generated text and $R$ denote the reference text.

\paragraph{ROUGE-1.}
ROUGE-1~\cite{lin2004rouge} measures unigram (1-gram) overlap between the generated and reference texts, serving as a proxy for lexical content coverage. 
Let $\text{Count}_{G}(w)$ and $\text{Count}_{R}(w)$ denote the frequency of unigram $w$ in $G$ and $R$, respectively. The unigram overlap is defined as:
\begin{equation}
\text{Overlap}(w) = \min(\text{Count}_{G}(w), \text{Count}_{R}(w)).
\end{equation}
The recall form of ROUGE-1 is computed as:
\begin{equation}
\text{ROUGE-1} =
\frac{\sum_{w \in R} \text{Overlap}(w)}
{\sum_{w \in R} \text{Count}_{R}(w)}.
\end{equation}
In practice, we report the F1-score to balance precision and recall:
\begin{equation}
\text{F1} =
\frac{2PR}{P + R},
\end{equation}
where $P$ and $R$ denote unigram precision and recall.

\paragraph{ROUGE-L.}
ROUGE-L~\cite{lin2004rouge} is based on the Longest Common Subsequence (LCS) between $G$ and $R$, capturing sentence-level structural similarity without requiring consecutive n-gram matches. Let $\text{LCS}(G,R)$ denote the length of the longest common subsequence. The recall and precision are defined as:
\begin{equation}
R_{LCS} =
\frac{\text{LCS}(G,R)}{|R|}, \quad
P_{LCS} =
\frac{\text{LCS}(G,R)}{|G|}.
\end{equation}
The ROUGE-L F-measure is:
\begin{equation}
\text{ROUGE-L} =
\frac{(1+\beta^2) R_{LCS} P_{LCS}}
{R_{LCS} + \beta^2 P_{LCS}},
\end{equation}
where $\beta$ is typically set to favor recall.

\paragraph{METEOR.}
METEOR (Metric for Evaluation of Translation with Explicit ORdering)~\cite{banerjee2005meteor} extends lexical overlap metrics by incorporating exact matches, stem matches, synonym matches, and a fragmentation penalty to account for word ordering. Let $m$ denote the number of matched unigrams between $G$ and $R$. Precision and recall are defined as:
\begin{equation}
P = \frac{m}{|G|}, \quad
R = \frac{m}{|R|}.
\end{equation}
METEOR computes a weighted harmonic mean emphasizing recall:
\begin{equation}
F_{\text{mean}} =
\frac{10PR}{R + 9P}.
\end{equation}
To penalize fragmented matches, a penalty term is introduced:
\begin{equation}
\text{Penalty} =
\gamma \left(\frac{c}{m}\right)^{\theta},
\end{equation}
where $c$ is the number of matched chunks and $\gamma, \theta$ are hyperparameters. The final METEOR score is:
\begin{equation}
\text{METEOR} =
F_{\text{mean}} \cdot (1 - \text{Penalty}).
\end{equation}

Overall, ROUGE-1 measures lexical overlap, ROUGE-L captures structural similarity via sequence alignment, and METEOR incorporates semantic matching and ordering penalties. Together, they provide complementary perspectives for evaluating personalized text generation quality.

\subsection{Training Details and Reproducibility}
\label{sec: training details and reproducibility}
For fair comparison, we conduct all experiments with 5 epochs with the AdamW optimizer. To avoid randomness, we set the generation temperature to be 0. For hyper-parameters, we tried different combinations and report the best results. Specifically, learning rate is chosen from $\{3e^{-4}, 1e^{-4},  5e^{-5}, 1e^{-5}, 5e^{-6}, 1e^{-6}\}$; component coefficient $s$ and $s'$ from $\{0.01, 0.1, 0.5, 1, 5, 10.0, 20.0, 50.0\}$. In sensitivity study, we investigate the influence of LoRA rank from $\{1,2,3,4,5,6,7,8,16,32,64,128\}$. All experiments were conducted on a single cluster node equipped with a Dell PowerEdge C6620
and NVIDIA H100 GPUs with 94 GB of memory.

\subsection{Llama2-7B Results}
Table~\ref{tab:llama2_results} reports the results using LLaMA-2-7b-chat as the backbone model. Both PLUME variants outperform prior personalized baselines. PLUME achieves slightly higher scores on several tasks, while PLUME-s attains the best overall balance between performance and parameter efficiency. Together with the Mistral-7B results in Table~\ref{tab:main_results}, these findings demonstrate the robustness of our method across different backbone models.

\section{More Analysis}
\subsection{Sensitivity on Number of Retrieved History interactions}
\begin{figure}
    \centering
    \includegraphics[width=0.6\linewidth]{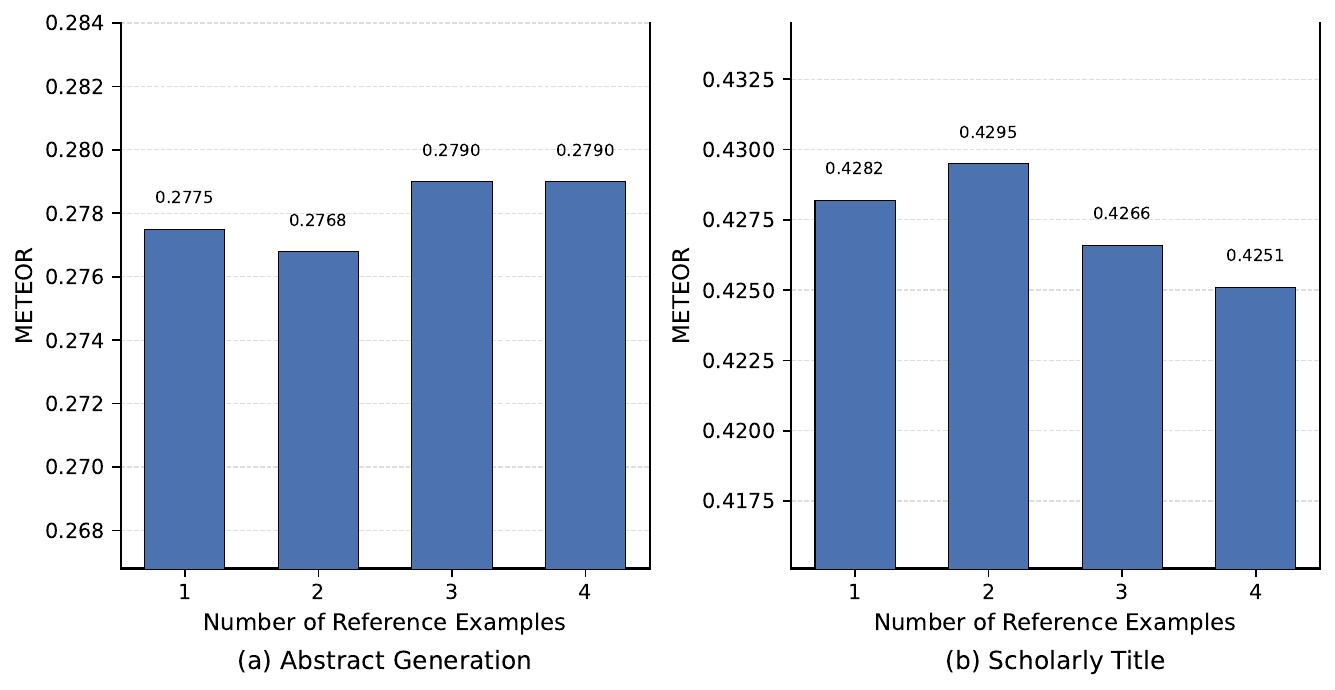}
    \caption{Performance variations on different numbers of retrieved interaction history.}
    \vspace{-3mm}
    \label{fig:Ablation_contrieve}
\end{figure}
From Figure ~\ref{fig:Ablation_contrieve}, we observe contrasting effects of increasing the number of in-prompt reference examples on personalized long-form content generation (e.g., Abstract Generation) versus short-form generation (e.g., Scholarly Title Generation) tasks. For Abstract Generation, adding examples yields small but consistent gains and then plateaus: performance nudges up from about 1 example to 3–4 examples, suggesting that extra stylistic cues help open-ended expansion without overwhelming the model. In contrast, for Scholarly Title Generation, quality peaks early and then declines: 1–2 examples give the best scores, while 3–4 examples slightly hurt, likely due to prompt dilution and competing keyword signals in a short, constrained output space. These findings indicate that long-form personalized generation benefits from more reference examples, whereas short-form generation achieves optimal performance with only 1–2 examples, providing practical guidance for designing in-context learning strategies tailored to different generation tasks.

\subsection{Performance Variation by Output Length}
\begin{figure}
    \centering
    \includegraphics[width=\linewidth]{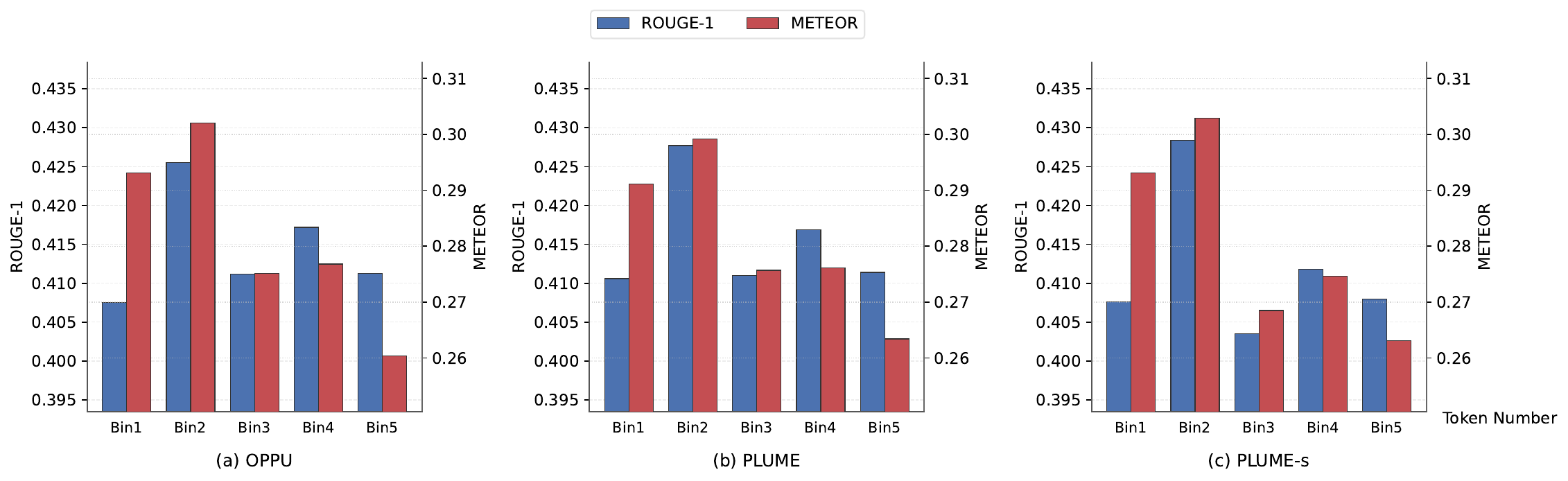}
    \caption{Abstract Generation performance across ground-truth length bins.
Bin 1: 79–125, Bin 2: 126–140, Bin 3: 141–155, Bin 4: 156–170, Bin 5: 171–245.
Panels show (a) OPPU, (b) PLUME, and (c) PLUME-s.}
    \vspace{-5mm}
    \label{fig:Group_Analysis}
\end{figure}


Figure ~\ref{fig:Group_Analysis} reports performance on the Abstract Generation task when users are grouped by the average number of ground-truth tokens. Both PLUME and PLUME-s consistently outperform OPPU across most length bins. A consistent trend appears across the OPPU, PLUME and PLUME-s models: ROUGE-1 and METEOR peak in the Bin 2 (token length 126–140) and then decline as ground-truth length increases. This trend highlights a critical characteristic in abstract generation tasks: there exists an optimal output length range for maintaining high summarization quality. When the ground-truth length exceeds a certain threshold—approximately 140 tokens—both lexical overlap (ROUGE-1) and semantic similarity (METEOR) degrade significantly, even though the models gradually increase their output length. From Table ~\ref{tab:gen_len_bins}, while the models generate progressively longer outputs with increasing bin size, they remain shorter and more conservative than the ground-truth abstracts. Longer ground-truth abstracts contain more entities and clause structure, so partial coverage and paraphrasing are penalized more strongly, revealing capacity limits of the per user LoRA adapters together with deterministic decoding. These results suggest using length aware decoding, explicit coverage objectives for salient terms and entities, and lightweight hierarchical planning to better match long ground-truth targets.

\section{Case Study}
To qualitatively illustrate the generation result of our approach over baseline approaches, we present a few representative examples from abstract generation and scholarly title generation tasks.

\vspace{10pt}
{%
  \setlength{\fboxsep}{6pt}%
  \colorbox[gray]{0.9}{%
    \parbox{\dimexpr\linewidth-2\fboxsep\relax}{\textbf{Prompt Template}}%
  }%
}
\vspace{-2pt}

\noindent
{%
    \setlength{\fboxsep}{6pt}
    \setlength{\fboxrule}{0.4pt}
    \doublebox{%
        \begin{minipage}{0.92\linewidth}
        \small
        You are an \textbf{academic researcher}.\\
        Your task is to generate an academic-style abstract that matches the author's writing style based on the paper's abstract.\\
        Here are reference examples (for style/tone ONLY; do not copy them): \texttt{<REFERENCE\_ABSTRACT>} is the abstract of \texttt{<REFERENCE\_TITLE>}.\\
        Generate a NEW abstract for the paper titled: \texttt{<TARGET\_TITLE>}\\
        \textbf{Constraints:}
        \begin{itemize}
          \item DO NOT copy any sentences from the reference abstracts.
          \item Length:150--250 words; formal, concise, and self-contained.
          \item Suggested structure: background/objective $\rightarrow$ method $\rightarrow$ data/pipeline $\rightarrow$ results/impact $\rightarrow$ (optional) deployment/cloud aspects.
          \item Write only the abstract text, without headings or extra commentary.
        \end{itemize}
        \end{minipage}%
    }%
}

\vspace{0.8em}

\definecolor{naturegray}{RGB}{249,249,249}
\definecolor{natureblue}{RGB}{236,243,250}
\definecolor{naturegold}{RGB}{255,252,235}
\definecolor{naturegreen}{RGB}{235,255,240}
\definecolor{naturered}{RGB}{255,240,240}
\sethlcolor{naturegold}

{\small
\setlength{\tabcolsep}{3pt}
\renewcommand{\arraystretch}{1.15}

\begin{xltabular}{\textwidth}{
  >{\raggedright\arraybackslash}p{0.18\textwidth}
  >{\raggedright\arraybackslash}X
}

\toprule
\textbf{Prompt} &
You are user \#009, and a academic researcher. Your task is to generate an academic-style abstract that matches the author's writing style based on the paper's abstract. Here are reference examples (for style/tone ONLY; do not copy them): ``The User Centric Smart Card Ownership Model (UCOM) provides an open and dynamic smart card environment enabling cardholders to request installation/deletion of an application to which they are entitled. As in this model, smart cards are not under the control of a centralised authority; hence, it is difficult for an application provider to ascertain their trustworthiness. At present, proposed secure channel protocols for the smart card environment do not provide adequate assurance required by the UCOM. In this paper, we explore the reasons behind their failure to meet the UCOM requirements and then propose a secure and trusted channel protocol that meets them. In addition, the proposed protocol is also suitable to GlobalPlatform's consumer-centric smart cards. A comparison of the proposed protocol with existing smart card and selected Internet protocols is provided. Then we analyse the protocol with the CasperFDR tool. Finally, we detail the implementation and the performance measurement.'' is the abstract of ``A Secure and Trusted Channel Protocol for the User Centric Smart Card Ownership Model.'' Generate a NEW abstract for the paper titled: ``Coopetitive architecture to support a dynamic and scalable NFC based mobile services architecture.'' Constraints: - DO NOT copy any sentences from the reference abstracts. - Length:150--250 words; formal, concise, and self-contained. - Suggested structure: background/objective $\rightarrow$ method $\rightarrow$ data/pipeline $\rightarrow$ results/impact $\rightarrow$ (optional) deployment/cloud aspects. Write only the abstract text, without headings or extra commentary. \#\#\# Response: \\
\midrule

\rowcolor[gray]{0.9}
\textbf{Version} & \textbf{Generated Abstract Output} \\
\midrule
\endfirsthead

\multicolumn{2}{c}{\tablename\ \thetable\ (continued)}\\
\toprule
\rowcolor[gray]{0.9}
\textbf{Version} & \textbf{Generated Abstract Output} \\
\midrule
\endhead

\midrule
\multicolumn{2}{r}{Continued on next page}\\
\endfoot

\bottomrule
\noalign{\vskip 6pt}
\caption{Case Study 1: NFC Coopetitive Architecture (Abstract Generation). Comparison across \textbf{Gold}, \textbf{Non-personalized LoRA}, \textbf{OPPU}, and \textbf{our PLUME-s}. Yellow highlights (\hl{phrases}) indicate generated content relevance while green highlights (\hlgreen{phrases}) indicate relevant ground truth.}
\label{tab:case_study_1}
\endlastfoot

\rowcolor{naturered}
\textbf{Gold (Reference)} &
\hlgreen{Near Field Communication (NFC)} has reinvigorated the multi-application smart card initiative. The NFC trials are relying on an extension of Issuer Centric Smart Card Model (ICOM) referred as Trusted Service Manager (TSM) architecture, which may create market segregation. Where the User Centric Smart Card Ownership Model (UCOM) takes an opposite approach of delegating the smart card ownership to its users. Therefore, to reconcile these two approaches we proposed the \hlgreen{Coopetitive Architecture for Smart Cards (CASC)} that avoids market segregation, increase revenue generation, and provide flexibility, robustness, and scalability. To support the CASC framework in this paper, we propose an \hlgreen{application installation protocol} that provides entity authentication, trust assurance and validation, mutual key and contractual-agreement generation. The protocol is compared with existing protocols on its performance, stated security, and operational goals. Furthermore, CasperFDR is used to provide a mechanical formal analysis of the protocol. \\
\midrule

\textbf{Non-personalized LoRA} &
\hl{We have designed a coopetitive architecture} to support a dynamic and scalable \hl{NFC based mobile service architecture}. Our architecture supports both cooperative and competitive business models. It enables service providers to offer location specific promotions to increase footfall and sales. \\
\midrule

\rowcolor{naturegray}
\textbf{OPPU} &
\hl{In this paper we propose a coopetitive architecture} for providing a generic and scalable framework to support \hl{Near Field Communication (NFC)} based mobile service deployments. The proposed architecture supports a coopetitive business model between service providers, as it enables them to collaborate while at the same time compete with each other. It provides a flexible and modular approach to deploy mobile services, where tiered level of service restrictions can be created. \hl{Furthermore, it provides a dynamic and future proof deployment strategy for mobile services} as it can easily accommodate new technologies, schemes or standards as they emerge. \\
\midrule

\rowcolor{natureblue}
\textbf{PLUME-s} &
\hl{In this paper, we propose a coopetitive architecture} that supports a dynamic and scalable \hl{Near Field Communication (NFC)} based mobile services architecture. The main objective of this research is to enable multiple service providers to cooperate and compete over a common platform while maintaining security and privacy of the sensitive information. \hl{To support this objective, we have designed and implemented a framework} that provides a flexible and robust approach to manage the life cycle of a mobile service. In addition, the framework provides a mechanism to ensure that only entitled mobile users access a particular service. \hl{Furthermore, the framework also provides a mechanism to prevent malicious entities from accessing the network and/or a particular service.} \\

\end{xltabular}
}

{\small
\setlength{\tabcolsep}{3pt}
\renewcommand{\arraystretch}{1.15}

\begin{xltabular}{\textwidth}{
  >{\raggedright\arraybackslash}p{0.18\textwidth}
  >{\raggedright\arraybackslash}X
}

\toprule
\textbf{Prompt} &
You are user \#0154, and an academic researcher. Your task is to generate a scholarly title that matches the author's writing style based on the paper's abstract. Here are reference examples (for style/tone ONLY; do not copy them): ``A cryptographic model for better information security.'' is the title for ``In this study, a novel cryptographic model that uses coupled map lattice is proposed for securing image. It incorporates mixing based on randomly generated secret key, sub-keys based substitution, confusion algorithm and coupled map lattice based diffusion process to enrich the security, sensitivity and robustness of the model.'' When mimicking the style, pay attention to academic tone, terminology, and title conventions. Never reuse exact phrases from the references. Now, following the style of the references, generate a title for this abstract: ``Face is a complex multidimensional visual model and developing a computational model for face recognition is difficult. The paper presents a methodology for face recognition based on information theory approach of coding and decoding the face image. Proposed methodology is connection of two stages - Feature extraction using principle component analysis and recognition using the feed forward back propagation Neural Network. The algorithm has been tested on 400 images (40 classes). A recognition score for test lot is calculated by considering almost all the variants of feature extraction. The proposed methods were tested on Olivetti and Oracle Research Laboratory (ORL) face database. Test results gave a recognition rate of 97.018\%.'' Write only the title, as if it were published in an academic journal. No explanations, no extra formatting. \#\#\# Response: \\
\midrule

\rowcolor[gray]{0.9}
\textbf{Version} & \textbf{Generated Title Output} \\
\midrule
\endfirsthead

\multicolumn{2}{c}{\tablename\ \thetable\ (continued)}\\
\toprule
\rowcolor[gray]{0.9}
\textbf{Version} & \textbf{Generated Title Output} \\
\midrule
\endhead

\midrule
\multicolumn{2}{r}{Continued on next page}\\
\endfoot

\bottomrule
\noalign{\vskip 6pt}
\caption{Case Study 2: Scholarly Title Generation. Comparison across \textbf{Gold}, \textbf{Non-personalized LoRA}, \textbf{OPPU}, and \textbf{our PLUME-s}. Yellow highlights (\hl{phrases}) indicate generated content relevance while green highlights (\hlgreen{phrases}) indicate relevant ground truth.}
\label{tab:case_study_2}
\endlastfoot

\rowcolor{naturered}
\textbf{Gold (Reference)} &
\hlgreen{Face recognition} using \hlgreen{principle component analysis}, \hlgreen{eigenface} and \hlgreen{neural network} \\
\midrule

\textbf{Non-personalized LoRA} &
Information theoretic approach for feature extraction and its application in \hl{face recognition} \\
\midrule

\rowcolor{naturegray}
\textbf{OPPU} &
Intuitional Analysis Based Approach for \hl{Recognizing Face} from Image \\
\midrule

\rowcolor{natureblue}
\textbf{PLUME-s} &
\hl{Recognition of face} using \hl{principal component analysis} and \hl{neural network} \\

\end{xltabular}
}

\normalsize
\smallskip
\clearpage
\noindent\textbf{Case Study Analysis.}
As shown in Table~\ref{tab:case_study_1} and Table~\ref{tab:case_study_2}, compared with OPPU, our model output demonstrates stronger fidelity to the target writing style and higher conceptual alignment with the gold abstract. It accurately highlights the core idea of a coopetitive architecture for Near Field Communication (NFC) based mobile services and elaborates the dual objective of enabling cooperation and competition among service providers within a secure and scalable framework. The output preserves the academic tone and structured exposition of the reference abstract, presenting a coherent progression from motivation and framework design to implementation and evaluation. Furthermore, it enhances the technical narrative by introducing details on life cycle management, access control, and security assurance, which contribute to the completeness and domain consistency of the abstract. In contrast, OPPU remains limited to a general architectural overview and omits key research components such as system objectives, evaluation context, and security mechanisms. This case demonstrates that our model effectively captures both stylistic and conceptual fidelity, producing outputs that are linguistically precise and technically well grounded in the target research domain. In the title generation case, the OPPU output deviates from academic conventions, using vague wording such as “Intuitional Analysis” and lacking precise technical terms. Our Personalized LoRA produces a clearer, domain-consistent title that mirrors the reference’s structure and terminology, showing better adaptation to scholarly style.

%% file: text/Related_work.tex
\section{Related Work}
\subsection{LoRA Based Fine Tuning}
LoRA (Low-Rank Adaptation)~\citep{hu2021lora} is one of the most widely adopted parameter-efficient fine-tuning (PEFT) techniques for adapting large language models (LLMs) with minimal trainable parameters. Instead of updating the full weight matrix, LoRA introduces a low-rank update $\Delta W = sAB$, where $A \in \mathbb{R}^{d \times r}$ and $B \in \mathbb{R}^{r \times k}$ with $r \ll \min({d, k})$, thus reducing the number of trainable parameters from $O(dk)$ to $O(r(d+k))$. Due to its effectiveness and efficiency, LoRA has become one of the most effective and practical approaches in personalized LLM adaptation tasks. Building upon this foundation, several variants have been developed to improve flexibility, stability, and efficiency. AdaLoRA~\citep{zhang2023adalora} dynamically allocates ranks through orthogonal regularization, DoRA~\citep{liu2024dora} decouples direction and magnitude learning to stabilize optimization, and QLoRA~\citep{dettmers2023qlora} integrates quantization for memory-efficient adaptation of large-scale models. A complementary line of work leverages singular value decomposition (SVD) for structured adaptation: PiSSA~\citep{meng2024pissa} initializes LoRA adapters with principal singular components for faster convergence, while MiLoRA~\citep{wang2024milora} and KASA~\citep{wang2024kasa} focus on minor singular directions to enhance generalization. Collectively, when personalization requires fine-tuning, LoRA and its variants serve as the most effective and widely adopted PEFT solutions, enabling scalable adaptation of LLMs to individual users while preserving efficiency and generalization.

\subsection{LLM personalization}
The problem of adapting large language models to individual users has been extensively studied, yielding a diverse landscape of personalization strategies. Typically, these paradigms could be classified into two categories: \emph{Prompt-based} and \emph{Adapter-based approaches}.

\noindent\textbf{Prompt-Based Approaches}
 Among the earliest and most lightweight approaches are prompt-based methods, which target to extract and encode user-specific information into handcrafted or learned prompts to guide model behavior without modifying LLM parameters~\cite{liu2025survey, liu2021pre}. In light of this, various prompting techniques, such as CoT and in-context learning, have been employed to provide a summarized or sampled user behavior history ~\cite{wang2023learning, kang2023llms}. For instance, Dai et al~\cite{dai2023uncovering} leverages sampled user purchase history to guide LLMs to make limited-size personalized item recommendations, while DPL~\cite{qiu2025measuring} improves personalization by encoding extracted inter-user comparisons into personalized prompts. However, such prompt-based approaches are fundamentally constrained by the model’s finite context window, and further suffer from the increasing difficulty of extracting effective information as user data grows. To mitigate these limitations, recent work has turned to \textit{retrieval-augmented personalized prompting}, which dynamically retrieves salient user records from a long-term memory to populate the prompt, obviating the need for exhaustive history inclusion~\cite{salemi2023lamp, qian2025memorag, sun2024persona}. As an example, Pearl~\cite{mysore2023pearl} leverages a retriever calibrated to the generation objective to select historical user-authored documents that properly enhance and augment the prompt. More recently, PRIME~\cite{zhang2025prime} proposes to enhance LLM personalization via episodic and semantic memory mechanisms, aiming to achieve retaining and updating memory for more efficient individual information retrieval. 

 While being conceptually simple and highly interpretable, prompt-based personalization relies on strong, explicit user signals in historical data ~\cite{tan2025aligning}. In content generation tasks, where personalization hinges on implicit factors such as stylistic preferences or individual personality traits, these approaches often struggle to reliably capture users' intent and preferences, leading to unstable and degraded performance.

\noindent\textbf{Adapter-Based Approaches}
Adapter-based approaches typically adapt PEFT methods and could be divided into two types, with the first category training all users with a shared model~\cite{zhu2024lifelong,li2024learning}; Existing work such as LM-P~\cite{wozniak2024personalized}, PLoRA\cite{zhang2024personalized} and MiLP~\cite{zhang2024personalized1} all fall in this research line. Recent advancements like iLoRA~\cite{kong2024customizing} and RecLoRA~\cite{zhu2024lifelong} have incorporated the Mixture of Experts (MoE) structure to model the diverse range of user preferences and behaviors. The second category of methods further enhances personalization performance by assigning each user a PEFT-trained model. For example, OPPU~\cite{tan2024democratizing} equips each user with a LoRA module and trains it on the user's individual data, achieving SOTA results on personalized classification and short content generation tasks on LaMP~\cite{salemi2023lamp} dataset. Further, CoPE~\cite{bu2025personalized} demonstrates the effectiveness of the strategy on long content generation tasks by augmenting training data with negative sampling together with contrastive loss. 

Despite its promising results, the main limitation of OPPU lies in its parameter growth. By assigning an individual PEFT module to each user, the total number of parameters increases rapidly with the number of users, resulting in considerable storage overhead. In addition, since each user typically has only a small amount of data, training a conventional PEFT module (such as LoRA) on such limited data often leads to overfitting and redundant parameters. Hence, reducing user-specific parameters is both urgent and essential for effective personalized LLMs.
This work first introduces this pivotal challenge, and proposes our solution--PLUME, which decomposes traditional LoRA into ultra-lightweight modules through a shared task-specific parameter space and a layer-wise sharing mechanism, achieving competitive personalization performance with minimal parameter redundancy.

\subsection{Details of Centered Kernel Alignment (CKA)}
\label{sec: cka_analysis}
We use linear Centered Kernel Alignment (CKA)~\cite{liu2025spectral} to measure representation similarity between LoRA parameters across layers. CKA provides a scale-invariant and rotation-invariant similarity measure, making it suitable for comparing learned adapter representations. 

Specifically, given two column-centered and vectorized LoRA weight matrices 
$X$ and $Y$, the linear CKA is defined as:
Given two column-centered vectorized LoRA matrices from different layers of the same user model, the linear CKA is defined as:
\begin{equation*}
\mathrm{CKA}(X,Y) = \frac{\|X^\top Y\|_F^2}{\|X^\top X\|_F \;\cdot\; \|Y^\top Y\|_F}.
\label{eq:cka}
\end{equation*}